\documentclass[11pt,a4paper]{article}
\usepackage[hyperref]{acl2017}
\usepackage{times}
\usepackage{url}
\usepackage{latexsym}
\usepackage{graphicx} % for eps graphics
\usepackage{url}
\usepackage{xspace}
\usepackage{lingmacros}
\usepackage{color,soul}
\usepackage[defaultlines=4,all]{nowidow}

\aclfinalcopy

\DeclareRobustCommand{\eg}{\textit{e.g.}\@\xspace}

\makeatletter
\DeclareRobustCommand{\etc}{% etc
    \@ifnextchar{.}%
        {etc}%
        {etc.\@\xspace}%
}
\makeatother

\DeclareRobustCommand{\ng}{$n$-gram\@\xspace}
\DeclareRobustCommand{\ngs}{$n$-grams\@\xspace}

\DeclareRobustCommand{\hatec}{\textsc{Hate}\@\xspace}
\DeclareRobustCommand{\offnc}{\textsc{Offensive}\@\xspace}
\DeclareRobustCommand{\okc}{\textsc{Ok}\@\xspace}

\title{Detecting Hate Speech in Social Media}

\author{Shervin Malmasi \\
  Harvard Medical School \\
 Boston, MA, United States \\
  {\tt smalmasi@bwh.harvard.edu} \\\And
  Marcos Zampieri \\
  University of Wolverhampton \\
  Wolverhampton, United Kingdom \\
  {\tt marcos.zampieri@uni-koeln.de} \\}

\date{}

\begin{document}
\maketitle
\begin{abstract}
\vspace{-0.3cm}
In this paper we examine methods to detect hate speech in social media, while distinguishing this from general profanity.
%something not been widely studied to date.
%
We aim to establish lexical baselines for this task by applying supervised classification methods 
%training a model on 
using a recently released dataset annotated for this purpose. As features, our system uses character \ngs, word \ngs and word skip-grams.
%
%We apply single classifiers as well as more advanced ensemble classifiers and stacked generalization, achieving the best result of $80\%$ accuracy for this 3-class classification task.
%
We obtain results of $78$\% accuracy in identifying posts across three classes.
Results demonstrate that the main challenge lies in discriminating profanity and hate speech from each other.
A number of directions for future work are discussed.
%A hate speech detection system such as the one presented in this paper can be integrated in a number of NLP applications to increase security and safety for Internet users of all ages.

%Analysis reveals that discriminating hate speech and profanity is not a simple task, which may require features that capture a deeper semantic understanding of the text not always possible with surface features.
%
%The variability of gold labels in the annotated data, due to differences in the subjective adjudications of the annotators, is also an issue. Other directions for future work are discussed.
\end{abstract}
%Remove 4.3, 7, 8, and 9.
%Baseline 
%Change the conclusion
\section{Introduction}
%\vspace{-0.1cm}

Research on safety and security in social media has grown substantially in the last decade. A particularly relevant aspect of this work is detecting and preventing the use of various forms of abusive language in blogs, micro-blogs, and social networks. A number of recent studies have been published on this issue such as the work by \newcite{xu2012learning} on identifying cyber-bullying, the detection of hate speech \cite{burnap2015cyber} which was the topic of a recent survey \cite{schmidt2017survey}, and the detection of racism \cite{tulkens2016dictionary} in user generated content.

The growing interest in this topic within the research community is evidenced by several related studies presented in Section \ref{sec:bg} and by two recent workshops: Text Analytics for Cybersecurity and Online Safety (TA-COS)\footnote{\url{http://www.ta-cos.org/home}} held in 2016 at LREC and Abusive Language Workshop (AWL)\footnote{\url{https://sites.google.com/site/abusivelanguageworkshop2017/}} held in 2017 at ACL.

In this paper we address the problem of hate speech detection using a dataset which contains English tweets annotated with three labels: (1)~hate speech (\hatec); (2)~offensive language but no hate speech (\offnc); and (3) no offensive content (\okc).  Most studies on abusive language so far \cite{burnap2015cyber,djuric2015hate,nobata2016abusive} have been modeled as binary classification with only one positive and one negative classes (\eg hate speech vs non-hate speech). As noted by \newcite{dinakar2011modeling}, systems trained on such data often rely on the frequency of offensive or non-socially acceptable words to distinguish between the two classes.  \newcite{dinakar2011modeling} stress that in some cases ``the lack of profanity or negativity [can] mislead the classifier".

Indeed, the presence of profane content does not in itself signify hate speech. 
General profanity is not necessarily targeted towards an individual and may be used for stylistic purposes or emphasis. On the other hand, hate speech may denigrate or threaten an individual or a group of people without the use of any profanities.

The main aim of this paper is to establish a lexical baseline for discriminating between hate speech and profanity on this standard dataset. The corpus used here provides us with an interesting opportunity to investigate how well a system can detect hate speech from other content that is generally profane. This baseline can be used to determine the difficulty of this task, and help highlight the most challenging aspects which must be addressed in future work. 

The rest of this paper is organized as follows. In Section 2 we briefly outline some previous work on abusive language detection. The data is presented in Section 3, along with a description of our computational approach, features, and evaluation methodology. Results are presented in Section 4, followed by a conclusion and future perspectives in Section \ref{sec:conclusion}.

\section{Related Work}
\label{sec:bg}
%\vspace{-0.1cm}

There have been several studies on computational methods to detect abusive language published in the last few years. One example is the work by \newcite{xu2012learning} who apply sentiment analysis to detect bullying in tweets and use Latent Dirichlet Allocation (LDA) topic models \cite{blei2003latent} to identify relevant topics in these texts.

A number of studies have been published on hate speech detection. As previously mentioned, to the best of our knowledge all of them rely on binary classification (\eg hate speech vs non-hate speech). Examples of such studies include the work by \newcite{kwok2013locate}, \newcite{djuric2015hate}, \newcite{burnap2015cyber}, and by \newcite{nobata2016abusive}.

Due to the availability of suitable corpora, the overwhelming majority of studies on abusive language, including ours, have used English data.  However, more recently a few studies have investigated abusive language detection in other languages. \newcite{mubarak2017} addresses abusive language detection on Arabic social media and \newcite{su2017} presents a system to detect and rephrase profanity in Chinese. Hate speech and abusive language datasets have been recently annotated for German \cite{ross2016measuring} and Slovene \cite{fiser2017} opening avenues for future work in languages other than English.

\section{Methods}

Next we present the Hate Speech Detection dataset used in our experiments. We applied a linear Support Vector Machine (SVM) classifier and used three groups of features extracted for these experiments: surface \ngs, word skip-grams, and Brown clusters. The classifier and features are described in more detail in Section \ref{sec:class} and Section \ref{sec:features} respectively.
Finally, Section \ref{sec:evaluation} discusses evaluation methods.

\subsection{Data}
%\vspace{-0.1cm}

In these experiments we use the aforementioned Hate Speech Detection dataset created by \newcite{davidson2017automated} and distributed via CrowdFlower.\footnote{\url{https://data.world/crowdflower/hate-speech-identification}} The dataset features $14{,}509$ English tweets annotated by a minimum of three annotators.

Individuals in charge of the annotation of this dataset were asked to annotate each tweet and categorize them into one of three classes:

\begin{enumerate}
\item (\hatec): contains hate speech;
\item (\offnc): contains offensive language but no hate speech;
\item (\okc): no offensive content at all.
\end{enumerate}

\noindent Each instance in this dataset contains the text of a tweet\footnote{ Each tweet is limited to a maximum of $140$ characters.} along with one of the three aforementioned labels. The distribution of the texts across the three classes is shown in Table~\ref{tab:data}.

\begin{table}[!ht]
\centering
\tabcolsep=0.25cm
\begin{tabular}{lr}
\hline
\textbf{Class} & \textbf{Texts} \\
\hline
\hatec	& $2{,}399$\\
\offnc	& $4{,}836$\\
\okc 	& $7{,}274$\\
\hline
Total	& $14{,}509$\\
\hline
\end{tabular}

\caption{The distribution of classes and tweets in the Hate Speech Detection dataset.}
\label{tab:data}
\vspace{-0.3cm}
\end{table}

\noindent All the texts are preprocessed to lowercase all tokens and to remove URLs and emojis.

\subsection{Classifier}
\label{sec:class}

We use a linear SVM to perform multi-class
classification in our experiments. We use the
LIBLINEAR\footnote{http://www.csie.ntu.edu.tw/\%7Ecjlin/liblinear/}
package \cite{LIBLINEAR} which has been shown to be very efficient for similar text classification tasks. For example, the LIBLINEAR SVM implementation has been demonstrated to be a very effective classifier for Native Language Identification \cite{malmasi-dras:2015:nli}, temporal text classification \cite{zampieri2016modeling}, and language variety identification \cite{zampieri2016computational}.

\subsection{Features}
\label{sec:features}

We use two groups of surface features in our experiments as follows:

\begin{itemize}
\item Surface \ngs: These are our most basic features, consisting of character $n$-grams (of~order $2$--$8$) and word $n$-grams (of~order $1$--$3$). All tokens are lowercased before extraction of \ngs; character \ngs are extracted across word boundaries.
\item Word Skip-grams: Similar to the above features, we also extract $1$-, $2$- and $3$-skip word bigrams. These features are were chosen to approximate longer distance dependencies between words, which would be hard to capture using bigrams alone.
\end{itemize}

\subsection{Evaluation}
\label{sec:evaluation}

To evaluate our methods we use $10$-fold cross-validation. For creating the folds, we employ stratified cross-validation aiming to ensure that the proportion of classes within each partition is equal \cite{kohavi:1995}. 

We report our results in terms of accuracy. The results obtained by our methods are compared against a majority class baseline and an oracle classifier.

The oracle takes the predictions by all the classifiers in Table~\ref{tab:results} into account. It assigns the correct class label for an instance if at least one of the the classifiers produces the correct label for that instance. This approach establishes the \textit{potential} or \textit{theoretical} upper limit performance for a given dataset. Similar analysis using oracle classifiers have been previously applied to estimate the theoretical upper bound of shared tasks datasets in Native Language Identification \cite{malmasi-tetreault-dras:2015} and similar language and language variety identification \cite{dslrec:2016}.

\section{Results}

We start by investigating  the efficacy of our features for this task. We fist train a single classifier, with each of them using a type of feature. Subsequently we also train a single model combining all of our features into single space. These are compared against the majority class baseline, as well as the oracle. The results of these experiments are listed in Table~\ref{tab:results}.

\begin{table}[!ht]
\renewcommand{\arraystretch}{1.0}
\center
\scalebox{1.0}{
\begin{tabular}{lr}
\hline
\textbf{Feature} & \textbf{Accuracy (\%)} \\
\hline
Majority Class Baseline				& $50.1$ \\
Oracle								& $91.6$ \\
\hline
% CHAR
Character bigrams		& $73.6$ \\
Character trigrams		& $77.2$ \\
Character $4$-grams		& $\mathbf{78.0}$ \\
Character $5$-grams		& $77.9$ \\
Character $6$-grams		& $77.2$ \\
Character $7$-grams		& $76.5$ \\
Character $8$-grams		& $75.8$ \\
%
%\rule{0pt}{3.5ex}%
\hline
Word unigrams			& $77.5$ \\
Word bigrams			& $73.8$ \\
Word trigrams			& $67.4$ \\
%
%\rule{0pt}{3.5ex}%
\hline
$1$-skip Word bigrams			& $74.0$ \\
$2$-skip Word bigrams			& $73.8$ \\
$3$-skip Word bigrams			& $73.9$ \\
\hline
%\rule{0pt}{3.5ex}%
All features combined 	& $77.5$ \\
\hline%\hline
\end{tabular}
}
\caption{Classification results under $10$-fold cross-validation.}
\label{tab:results}
%\vspace{-0.4cm}
\end{table}

The majority class baseline is quite high due to the class imbalance in the data. The oracle achieves an accuracy of $91.6\%$, showing that none of our features are able to correctly classify a substantial portion of our samples.

We note that character \ngs perform well here, with $4$-grams achieving the best performance of all features. Word unigrams also perform well, while performance degrades with bigrams, trigrams and skip-grams.
However, the skip-grams may be capturing longer distance dependencies which provide complementary information to the other feature types.
In tasks relying on stylistic information, it has been shown that skip-grams capture information that is very similar to syntactic dependencies \cite[\S5]{malmasi-cahill:2015}.

Finally, the combination of all features does not achieve the performance of a character $4$-grams model and  causes a large dimensionality increase, with a total of $5.5$ million features.
It is not clear if this model is able to correctly capture the diverse information provided by the three feature types since we include more character \ng models than word-based ones.

\begin{figure}[!ht]
\centering
\includegraphics[trim=28 5 35 30,clip,width=0.47\textwidth]{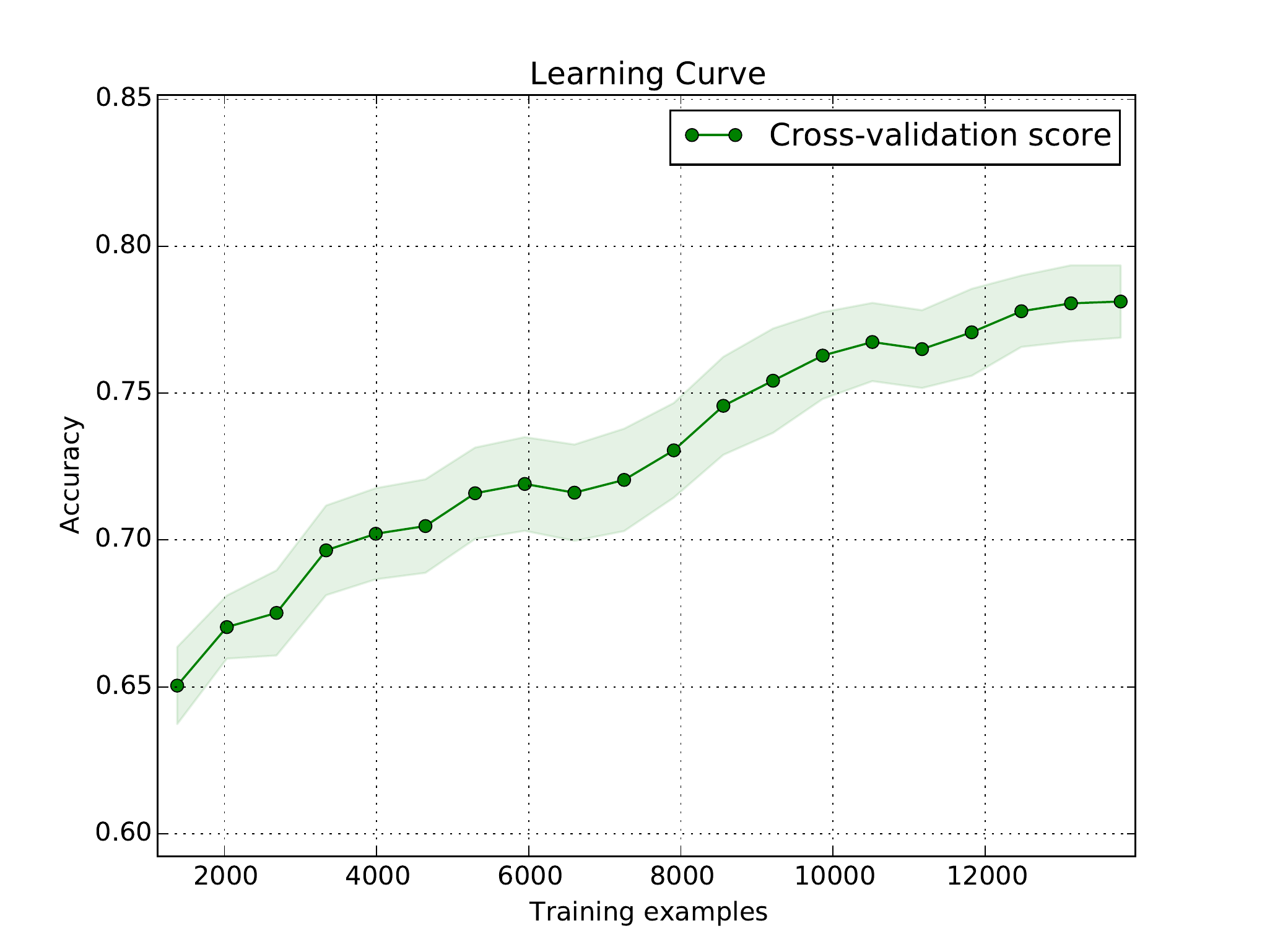}
\vspace{-0.5cm}
\caption{Learning curve for a character $4$-gram model, with standard deviation highlighted. Accuracy does not plateau with the maximal data size.}
\label{fig:learning-curve}
%\vspace{-0.3cm}
\end{figure}

% LEARNING CURVE
Next we analyze the rate of learning for these features. A learning curve for the classifier that yielded the best performance overall, character 4-grams, is shown in Figure \ref{fig:learning-curve}.

\noindent We observe that accuracy increased continuously as the amount of training instances increased, and the standard deviation of the results between the cross-validation folds decreased. This suggests that the use of more training data is likely to provide even higher accuracy. It should be noted, however, that accuracy increases at a much slower rate after $15,000$ training instances.

\begin{figure*}[!t]
\begin{center}
\includegraphics[width=0.600\textwidth]{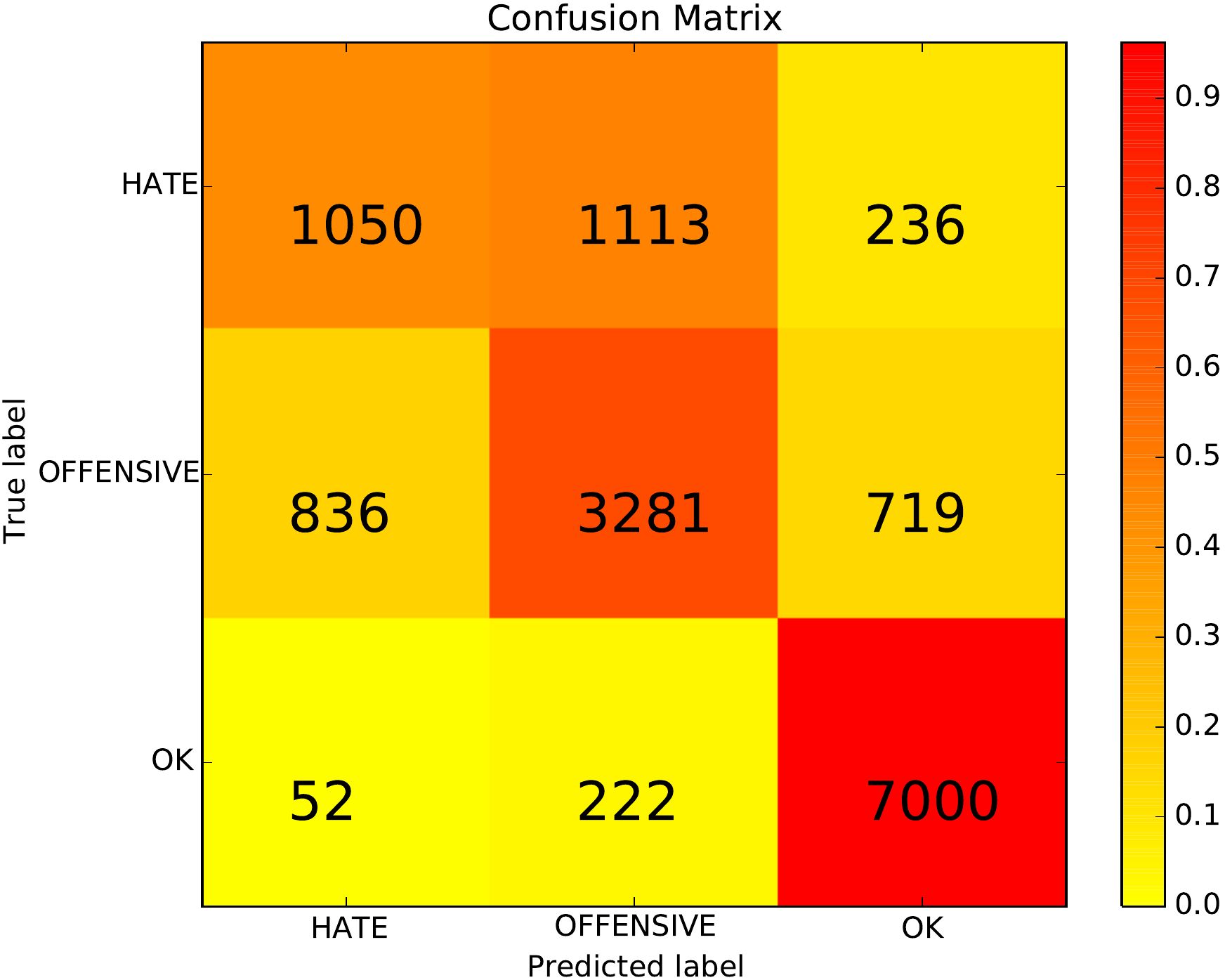}
\end{center}
\caption{Confusion matrix of the character 4-gram model for our $3$ classes. The heatmap represents the proportion of correctly classified examples in each class (this is normalized as the data distribution is imbalanced). The raw numbers are also reported within each cell. We note that the \hatec class is the hardest to classify and is highly confused with the \offnc class.}
\label{fig:confm}
\end{figure*}

\pagebreak

\noindent Finally, we also examine a confusion matrix for the character $4$-gram model, as shown in Figure~\ref{fig:confm}.
This demonstrates that the greatest degree of confusion lies between hate speech and generally offensive material, with hate speech more frequently being confused for offensive content. A substantial amount of offensive content is also misclassified as being non-offensive.
The non-offensive class achieves the best result, with the vast majority of samples being correctly classified.

\section{Conclusion}
\label{sec:conclusion}

In this paper we applied text classification methods to distinguish between hate speech, profanity, and other texts. We applied standard lexical features and a linear SVM classifier to establish a baseline for this task. The best result was obtained by a character 4-gram model achieving $78\%$ accuracy. The results presented in this paper showed that distinguishing profanity from hate speech is a very challenging task.

This was to the best of our knowledge one of the first experiments to detect hate speech on social media in a scenario including non-hate speech profanity. Previous work so far (e.g. \newcite{burnap2015cyber} and \newcite{djuric2015hate}) dealt with the distinction between hate speech and socially acceptable texts in a binary classification setting. In binary classification, \newcite{dinakar2011modeling} note that the frequency of offensive words helps classifiers to distinguish between hate speech and socially acceptable texts. 

We see a few directions in which this work could be expanded such as the use of more robust ensemble classifiers, a linguistic analysis of the most informative features, and error analysis of the misclassified instances. These aspects are presented in more detail in the next section.

\subsection{Future Work}
\label{sec:futurework}

In future work we would like to investigate the performance of classifier ensembles and meta-learning for this task. Previous work has applied these techniques to a number of comparable text classification tasks, achieving success in competitive shared tasks. Examples of recent applications include automatic triage of posts in mental health forums \cite{malmasi2016predicting}, detection of lexical complexity \cite{malmasi2016ltg}, Native Language Identification \cite{malmasi:2017:nlisg}, and dialect identification \cite{malmasi2017german}.

Another direction to pursue is the careful analysis of the most informative features for each class in this dataset. Our initial exploitation of the most informative words unigrams and bigrams suggests that coarse and obscene words are very informative for both \hatec and \offnc words which confuses the classifiers. For \hatec we observed a prominence of words targeting ethnic and social groups. Finally, an interesting outcome that should be investigated in more detail is that many of the most informative bigrams for the \okc feature grammatical words. A more detailed analysis of these features could lead to more robust feature engineering methods.

An error analysis could also help us better understand the challenges in this task. This could be used to provide insights about the classifiers' performance as well as any underlying issues with the annotation of the Hate Speech Detection dataset which, as pointed out by \newcite{ross2016measuring}, is far from trivial. Figure \ref{fig:confm} confirms that, as expected, most confusion occurs between \hatec and \offnc texts. However, we also note that a substantial amount of offensive content is misclassified as being non-offensive. The aforementioned error analysis can provide insights about this.

\section*{Acknowledgments}

We would like to thank the anonymous RANLP reviewers who provided us valuable feedback to increase the quality of this paper.

We further thank the developers and the annotators who worked on the Hate Speech Dataset for making this important resource available.

%\printbibliography
\bibliographystyle{acl_natbib}

\bibliography{ensemble,dsl,wordrep,profanity}

\begin{thebibliography}{}
\expandafter\ifx\csname natexlab\endcsname\relax\def\natexlab#1{#1}\fi

\bibitem[{Blei et~al.(2003)Blei, Ng, and Jordan}]{blei2003latent}
David~M Blei, Andrew~Y Ng, and Michael~I Jordan. 2003.
\newblock {Latent Dirichlet Allocation}.
\newblock {\em Journal of machine Learning research\/} 3(Jan):993--1022.

\bibitem[{Burnap and Williams(2015)}]{burnap2015cyber}
Pete Burnap and Matthew~L Williams. 2015.
\newblock Cyber hate speech on twitter: An application of machine
  classification and statistical modeling for policy and decision making.
\newblock {\em Policy \& Internet\/} 7(2):223--242.

\bibitem[{Davidson et~al.(2017)Davidson, Warmsley, Macy, and
  Weber}]{davidson2017automated}
Thomas Davidson, Dana Warmsley, Michael Macy, and Ingmar Weber. 2017.
\newblock {Automated Hate Speech Detection and the Problem of Offensive
  Language}.
\newblock In {\em Proceedings of ICWSM\/}.

\bibitem[{Dinakar et~al.(2011)Dinakar, Reichart, and
  Lieberman}]{dinakar2011modeling}
Karthik Dinakar, Roi Reichart, and Henry Lieberman. 2011.
\newblock Modeling the detection of textual cyberbullying.
\newblock In {\em The Social Mobile Web\/}. pages 11--17.

\bibitem[{Djuric et~al.(2015)Djuric, Zhou, Morris, Grbovic, Radosavljevic, and
  Bhamidipati}]{djuric2015hate}
Nemanja Djuric, Jing Zhou, Robin Morris, Mihajlo Grbovic, Vladan Radosavljevic,
  and Narayan Bhamidipati. 2015.
\newblock Hate speech detection with comment embeddings.
\newblock In {\em Proceedings of the 24th International Conference on World
  Wide Web Companion\/}. International World Wide Web Conferences Steering
  Committee, pages 29--30.

\bibitem[{Fan et~al.(2008)Fan, Chang, Hsieh, Wang, and Lin}]{LIBLINEAR}
Rong-En Fan, Kai-Wei Chang, Cho-Jui Hsieh, Xiang-Rui Wang, and Chih-Jen Lin.
  2008.
\newblock {{LIBLINEAR}: A Library for Large Linear Classification}.
\newblock {\em Journal of Machine Learning Research\/} 9:1871--1874.

\bibitem[{Fi\v{s}er et~al.(2017)Fi\v{s}er, Erjavec, and
  Ljube\v{s}i\'{c}}]{fiser2017}
Darja Fi\v{s}er, Toma\v{z} Erjavec, and Nikola Ljube\v{s}i\'{c}. 2017.
\newblock {Legal Framework, Dataset and Annotation Schema for Socially
  Unacceptable On-line Discourse Practices in Slovene}.
\newblock In {\em Proceedings of the Workshop Workshop on Abusive Language
  Online (ALW)\/}. Vancouver, Canada.

\bibitem[{Goutte et~al.(2016)Goutte, L{\'e}ger, Malmasi, and
  Zampieri}]{dslrec:2016}
Cyril Goutte, Serge L{\'e}ger, Shervin Malmasi, and Marcos Zampieri. 2016.
\newblock {Discriminating Similar Languages: Evaluations and Explorations}.
\newblock In {\em Proceedings of Language Resources and Evaluation (LREC)\/}.
  Portoroz, Slovenia.

\bibitem[{Kohavi(1995)}]{kohavi:1995}
Ron Kohavi. 1995.
\newblock {A study of cross-validation and bootstrap for accuracy estimation
  and model selection}.
\newblock In {\em IJCAI\/}. volume~14, pages 1137--1145.

\bibitem[{Kwok and Wang(2013)}]{kwok2013locate}
Irene Kwok and Yuzhou Wang. 2013.
\newblock Locate the hate: Detecting tweets against blacks.
\newblock In {\em Twenty-Seventh AAAI Conference on Artificial Intelligence\/}.

\bibitem[{Malmasi and Cahill(2015)}]{malmasi-cahill:2015}
Shervin Malmasi and Aoife Cahill. 2015.
\newblock {Measuring Feature Diversity in Native Language Identification}.
\newblock In {\em Proceedings of the Tenth Workshop on Innovative Use of NLP
  for Building Educational Applications\/}. Association for Computational
  Linguistics, Denver, Colorado.

\bibitem[{Malmasi and Dras(2015)}]{malmasi-dras:2015:nli}
Shervin Malmasi and Mark Dras. 2015.
\newblock {Large-scale Native Language Identification with Cross-Corpus
  Evaluation}.
\newblock In {\em Proceedings of NAACL-HLT 2015\/}. Association for
  Computational Linguistics, Denver, Colorado.

\bibitem[{Malmasi and Dras(2017)}]{malmasi:2017:nlisg}
Shervin Malmasi and Mark Dras. 2017.
\newblock {Native Language Identification using Stacked Generalization}.
\newblock {\em arXiv preprint arXiv:1703.06541\/} .

\bibitem[{Malmasi et~al.(2016{\natexlab{a}})Malmasi, Dras, and
  Zampieri}]{malmasi2016ltg}
Shervin Malmasi, Mark Dras, and Marcos Zampieri. 2016{\natexlab{a}}.
\newblock Ltg at semeval-2016 task 11: Complex word identification with
  classifier ensembles.
\newblock In {\em Proceedings of SemEval\/}.

\bibitem[{Malmasi et~al.(2015)Malmasi, Tetreault, and
  Dras}]{malmasi-tetreault-dras:2015}
Shervin Malmasi, Joel Tetreault, and Mark Dras. 2015.
\newblock {Oracle and Human Baselines for Native Language Identification}.
\newblock In {\em Proceedings of the Tenth Workshop on Innovative Use of NLP
  for Building Educational Applications\/}. Association for Computational
  Linguistics, Denver, Colorado.

\bibitem[{Malmasi and Zampieri(2017)}]{malmasi2017german}
Shervin Malmasi and Marcos Zampieri. 2017.
\newblock {German Dialect Identification in Interview Transcriptions}.
\newblock In {\em Proceedings of the Workshop on NLP for Similar Languages,
  Varieties and Dialects (VarDial)\/}.

\bibitem[{Malmasi et~al.(2016{\natexlab{b}})Malmasi, Zampieri, and
  Dras}]{malmasi2016predicting}
Shervin Malmasi, Marcos Zampieri, and Mark Dras. 2016{\natexlab{b}}.
\newblock {Predicting Post Severity in Mental Health Forums}.
\newblock In {\em Proceedings of the Workshop on Computational Linguistics and
  Clinical Psychology (CLPsych)\/}.

\bibitem[{Mubarak et~al.(2017)Mubarak, Kareem, and Walid}]{mubarak2017}
Hamdy Mubarak, Darwish Kareem, and Magdy Walid. 2017.
\newblock {Abusive Language Detection on Arabic Social Media}.
\newblock In {\em Proceedings of the Workshop Workshop on Abusive Language
  Online (ALW)\/}. Vancouver, Canada.

\bibitem[{Nobata et~al.(2016)Nobata, Tetreault, Thomas, Mehdad, and
  Chang}]{nobata2016abusive}
Chikashi Nobata, Joel Tetreault, Achint Thomas, Yashar Mehdad, and Yi~Chang.
  2016.
\newblock {Abusive Language Detection in Online User Content}.
\newblock In {\em Proceedings of the 25th International Conference on World
  Wide Web\/}. International World Wide Web Conferences Steering Committee,
  pages 145--153.

\bibitem[{Ross et~al.(2016)Ross, Rist, Carbonell, Cabrera, Kurowsky, and
  Wojatzki}]{ross2016measuring}
Bj{\"o}rn Ross, Michael Rist, Guillermo Carbonell, Benjamin Cabrera, Nils
  Kurowsky, and Michael Wojatzki. 2016.
\newblock {Measuring the Reliability of Hate Speech Annotations: The Case of
  the European Refugee Crisis}.
\newblock In {\em Proceedings of the Workshop on Natural Language Processing
  for Computer-Mediated Communication (NLP4CMC)\/}. Bochum, Germany.

\bibitem[{Schmidt and Wiegand(2017)}]{schmidt2017survey}
Anna Schmidt and Michael Wiegand. 2017.
\newblock {A Survey on Hate Speech Detection Using Natural Language
  Processing}.
\newblock In {\em Proceedings of the Fifth International Workshop on Natural
  Language Processing for Social Media. Association for Computational
  Linguistics\/}. Valencia, Spain, pages 1--10.

\bibitem[{Su et~al.(2017)Su, Huang, Chang, and Lin}]{su2017}
Huei-Po Su, Chen-Jie Huang, Hao-Tsung Chang, and Chuan-Jie Lin. 2017.
\newblock {Rephrasing Profanity in Chinese Text}.
\newblock In {\em Proceedings of the Workshop Workshop on Abusive Language
  Online (ALW)\/}. Vancouver, Canada.

\bibitem[{Tulkens et~al.(2016)Tulkens, Hilte, Lodewyckx, Verhoeven, and
  Daelemans}]{tulkens2016dictionary}
St{\'e}phan Tulkens, Lisa Hilte, Elise Lodewyckx, Ben Verhoeven, and Walter
  Daelemans. 2016.
\newblock {A Dictionary-based Approach to Racism Detection in Dutch Social
  Media}.
\newblock In {\em Proceedings of the Workshop Text Analytics for Cybersecurity
  and Online Safety (TA-COS)\/}. Portoroz, Slovenia.

\bibitem[{Xu et~al.(2012)Xu, Jun, Zhu, and Bellmore}]{xu2012learning}
Jun-Ming Xu, Kwang-Sung Jun, Xiaojin Zhu, and Amy Bellmore. 2012.
\newblock Learning from bullying traces in social media.
\newblock In {\em Proceedings of the 2012 conference of the North American
  chapter of the association for computational linguistics: Human language
  technologies\/}. Association for Computational Linguistics, pages 656--666.

\bibitem[{Zampieri et~al.(2016{\natexlab{a}})Zampieri, Malmasi, and
  Dras}]{zampieri2016modeling}
Marcos Zampieri, Shervin Malmasi, and Mark Dras. 2016{\natexlab{a}}.
\newblock {Modeling language change in historical corpora: the case of
  Portuguese}.
\newblock In {\em Proceedings of Language Resources and Evaluation (LREC)\/}.
  Portoroz, Slovenia.

\bibitem[{Zampieri et~al.(2016{\natexlab{b}})Zampieri, Malmasi, Sulea, and
  Dinu}]{zampieri2016computational}
Marcos Zampieri, Shervin Malmasi, Octavia-Maria Sulea, and Liviu~P Dinu.
  2016{\natexlab{b}}.
\newblock {A Computational Approach to the Study of Portuguese Newspapers
  Published in Macau}.
\newblock In {\em Proceedings of Workshop on Natural Language Processing Meets
  Journalism (NLPMJ)\/}. pages 47--51.

\end{thebibliography}

\end{document}